\documentclass[runningheads]{llncs}

\usepackage[T1]{fontenc}
\usepackage[utf8]{inputenc}
\usepackage{graphicx}
\usepackage{amsmath,amssymb}
\usepackage{booktabs}
\usepackage{multirow}
\usepackage{algorithm}
\usepackage[noend]{algpseudocode}
\usepackage{url}

\newcommand{\ldt}{\textsc{ldt}}

\makeatletter
\renewcommand\section{\@startsection{section}{1}{\z@}%
	{0.8ex \@plus 0.8ex \@minus 0.8ex}%
	{0.8ex \@plus 0.8ex}%
	{\normalfont\large\bfseries}}
\renewcommand\subsection{\@startsection{subsection}{2}{\z@}%
	{0.7ex \@plus 0.7ex \@minus 0.7ex}%
	{0.7ex \@plus 0.7ex}%
	{\normalfont\normalsize\bfseries}}
\renewcommand\subsubsection{\@startsection{subsubsection}{3}{\z@}%
	{0.7ex \@plus 0.7ex \@minus 0.7ex}%
	{0.7ex \@plus 0.7ex}%
	{\normalfont\normalsize\itshape}}
\makeatother

\begin{document}
	
	\title{An Incremental Knowledge-Graph Learner Digital Twin for Evidence-Grounded Early Warning}
	\titlerunning{A Knowledge-Graph Learner Digital Twin}
	
	\author{
		Ng\d{o}c Luy\d{\^{e}}n L\^e\inst{1,2}\orcidID{0000-0002-2043-0679}
		\and
		Marie-H\'el\`ene Abel\inst{2}\orcidID{0000-0003-1812-6763}
		\and
		Bertrand Laforge\inst{3}\orcidID{0000-0002-4209-4194}
	}
	
	\authorrunning{N. L. L\^e et al.}
	
	\institute{
		Gamaizer, 93340 Le Raincy, France
		\and
		Universit\'e de technologie de Compi\`egne, CNRS, Heudiasyc
		(Heuristics and Diagnosis of Complex Systems),
		CS 60319, 60203 Compi\`egne Cedex, France\\
		\email{\{ngoc-luyen.le,marie-helene.abel\}@hds.utc.fr}
		\and
		Sorbonne Universit\'e, CNRS UMR 7585, LPNHE
		(Laboratoire de Physique Nucl\'eaire et des Hautes \'Energies),
		75252 Paris Cedex 05, France\\
		\email{laforge@lpnhe.in2p3.fr}
	}

	\maketitle
	\vspace{-0.7cm}
	
	\begin{abstract}
Early-warning models in learning analytics can identify learners at risk of failure or withdrawal, but the resulting alerts are often only weakly connected to inspectable learning evidence. A risk score may indicate that intervention is needed without showing which traces, assessment events, or temporal patterns justify the alert. This paper introduces an incremental \textit{Learner Digital Twin} for evidence-grounded early warning, implemented as a lightweight knowledge graph. The twin is updated at temporal checkpoints by replaying learning traces and assessment records, and it serves both as a feature-extraction substrate and as an explanation-grounding substrate. Predictive features are derived through graph-mediated queries, while features, attributions, alerts, evidence items, and raw records are represented as queryable graph entities. The framework is instantiated and evaluated on a public learning-analytics benchmark under a temporal protocol. The results show that graph-mediated feature extraction preserves the predictive behavior of equivalent raw-table features across model families and checkpoints. Generated alerts are materialized as evidence chains linking risk decisions to feature attributions, evidence items, and raw platform records. The study shows that early-warning prediction can be routed through a semantic and provenance-aware representation while preserving predictive performance and adding traceability, verifiability, and evidence-grounded explanation.

	\end{abstract}
	
\section{Introduction}
	\label{sec:intro}
	
Web-based learning platforms continuously record learner activity, including resource accesses, submissions, scores, registration events, assessment results, and withdrawals. Learning analytics has built on these traces to develop early-warning systems that estimate, before the end of a course, which learners may be at risk of failing or dropping out~\cite{akccapinar2019using,herodotou2019}. Such systems can support timely pedagogical intervention, but their usefulness depends on more than predictive accuracy. For an alert to be actionable, instructors also need to understand which learning traces, assessment events, or temporal patterns justify the risk signal.

The predictive dimension of early-warning learning analytics is now relatively well served. Tree-based ensembles, gradient-boosting methods, and neural architectures can achieve strong discrimination when trained on interaction and assessment features~\cite{kuzilek2017,waheed2020}. However, the explanatory status of the resulting alerts remains limited. In a typical pipeline, raw platform logs are transformed into tabular features, a predictive model assigns a risk score, and a post-hoc attribution method such as SHAP~\cite{lundberg2017} may indicate that an engineered variable, for example \textit{total\_clicks}, contributed to the prediction. Such feature-level explanations are useful for model inspection, but they often stop short of what instructors need for pedagogical interpretation and action. An instructor-facing alert should not only indicate that a learner is at risk; it should also make explicit which learning evidence supports this claim. A feature attribution names an engineered quantity, but it does not necessarily identify the concrete learning events behind it: which interactions decreased, which assessment was missed, which raw records substantiate the alert, and whether the explanation can be queried, traced, or contested. This creates a gap between predictive performance and evidence-grounded interpretability, a limitation emphasized in explainable learning analytics and AI-in-education research~\cite{khosravi2022}.

We address the following problem: given a stream of timestamped learning traces and assessment records, how can an early-warning system produce risk alerts that are not only predictive, but also evidence-grounded, queryable, and traceable? To this end, we introduce an \emph{incremental Learner Digital Twin} (\ldt{}) implemented as a lightweight knowledge graph (KG). The twin is updated at successive temporal checkpoints by replaying learning traces and assessment data in chronological order, thereby simulating the recurrent synchronization that a \ldt{} would perform in a live learning platform. The central design principle is that the graph is not a post-hoc explanatory layer added after prediction; instead, it acts as the operational substrate for feature construction, prediction, attribution representation, and evidence verification. At each checkpoint, learner feature vectors are extracted from the twin using a graph-mediated query catalogue and passed to an early-warning classifier. The resulting prediction features, feature attributions, generated alerts, and evidence items are then materialized back into the graph, with each evidence item linked to raw learning records following a provenance-aware modeling perspective inspired by PROV-O~\cite{lebo2013prov}. Graph integrity is checked using SHACL validation~\cite{knublauch2017shacl}, so that explanation chains can be verified as graph properties. The framework is later instantiated and evaluated on the Open University Learning Analytics Dataset (OULAD)~\cite{kuzilek2017}.

The objective is not to improve predictive accuracy, but to show that early-warning prediction can be routed through a semantic and provenance-aware representation while preserving predictive performance and making alerts evidence-grounded, queryable, and traceable. The paper makes three contributions: it introduces an incremental knowledge-graph \ldt{} synchronized at temporal checkpoints from learning traces and assessment data; it represents predictive features, feature attributions, alerts, evidence items, and raw records as first-class graph entities; and it empirically verifies that graph-mediated feature extraction preserves the behavior of an equivalent raw-table pipeline while enabling SHACL-validated evidence grounding and traceability.

The remainder of this paper is organized as follows. Section~\ref{sec:rw} reviews related work on early warning, explainable learning analytics, learner digital twins, and semantic provenance. Section~\ref{sec:method} presents the knowledge-graph \ldt{}. Section~\ref{sec:exp} describes the experimental setup and results. Section~\ref{sec:discussion} discusses implications and limitations. Section~\ref{sec:conclusion} concludes the paper.
	
\section{Related Work}
\label{sec:rw}
	
We review evidence-grounded early warning in learning analytics. We focus on three areas: early-warning and explainable learning analytics, learner modeling and digital twins, and knowledge graphs for provenance and traceability.

%Early-warning systems are among the most established applications of learning analytics. They use traces of learner activity, assessment behavior, registration information, and other platform data to identify learners who may fail, withdraw, or require support before the end of a course~\cite{arnold2012course}. Operational systems such as Course Signals~\cite{arnold2012course} and OU Analyse~\cite{hlosta2017,herodotou2019empowering} illustrate how predictive learning analytics has been deployed as institutional decision support. 
Early-warning systems are among the most established applications of
learning analytics. They use traces of learner activity, assessment behavior,
registration information, and other platform data to identify learners who
may fail, withdraw, or require support before the end of a course.
Operational systems such as Course Signals~\cite{arnold2012course} and OU
Analyse~\cite{hlosta2017} illustrate how predictive learning analytics has
been deployed as institutional decision support.
In parallel, public benchmark datasets have enabled reproducible experimentation
on early risk prediction using activity and assessment
traces~\cite{kuzilek2017}.
Related work has also examined leakage-excluded early-warning
	prediction under strict temporal constraints~\cite{le2026can} and
	semantics-constrained counterfactual recourse for educational decision
	support~\cite{le2026student}.
Most work in this area has focused on predictive performance, timeliness,
and deployment feasibility.
	
Explainability is particularly important in learning analytics because predictions may inform pedagogical decisions, learner support, or institutional interventions. Post-hoc explanation methods such as LIME and SHAP are commonly used because they attribute individual predictions to input features without requiring a fully transparent model architecture~\cite{ribeiro2016,lundberg2017}. However, a feature-level attribution remains incomplete if the instructor cannot inspect which resources were accessed, which assessment was missed, or which raw records substantiate the feature. The issue is therefore not only whether an attribution is locally faithful to the model, but whether the attributed feature is grounded in interpretable educational evidence.

Learner modeling has long addressed the representation of learner states. Overlay models, open learner models, and adaptive educational systems represent learners' knowledge, progress, preferences, goals, and interaction histories~\cite{bull2007}. Ontology-based learner models further contribute semantic structure and interoperability by representing learners, contexts, resources, and educational situations through explicit conceptual models~\cite{saleh2018sis}. However, many learner models are built from a given snapshot of learner data or are used primarily as profiles for adaptation and visualization. They are less often operationalized as time-indexed representations that are recurrently synchronized with platform traces and directly connected to predictive alert generation.
	
%Learner modeling has long addressed the representation of learner states. Overlay models, open learner models, and adaptive educational systems represent learners' knowledge, progress, preferences, goals, and interaction histories~\cite{bull2007,brusilovsky2007user}. Ontology-based learner models further contribute semantic structure and interoperability by representing learners, contexts, resources, and educational situations through explicit conceptual models~\cite{saleh2018sis,li2021collab}. However, many learner models are built from a given snapshot of learner data or are used primarily as profiles for adaptation and visualization. They are less often operationalized as time-indexed representations that are recurrently synchronized with platform traces and directly connected to predictive alert generation.
	
The notion of a digital twin originated in product lifecycle management, where it refers to a virtual counterpart of a physical asset connected to the real system through data flows~\cite{grieves2017}. Work in smart manufacturing emphasizes that a digital twin is not merely a static model, but a data-driven representation updated from observations and used to support analysis, simulation, and decision-making~\cite{tao2018digital,tao2019dt}. When transposed to education, the twinned entity may be a learner, a course, or a learning environment~\cite{habarurema2025enhancing}. In this paper, we use the term in a restricted operational sense: a checkpoint-indexed learner representation synchronized by replayed platform observations and used to support prediction and evidence-grounded explanation.
	
Ontologies and semantic web technologies have been used to structure learner information, educational resources, competencies, contexts, and learning activities. Early standards such as IMS LIP provided foundations for learner profile interoperability~\cite{imsLIP2001}. Educational knowledge graphs and linked learning resources have also shown how semantic representations can support search, recommendation, integration, and reasoning over educational data~\cite{chen2018knowedu,dietze2015inked}. These approaches demonstrate the value of semantic modeling, but they generally focus on representing learners, resources, activities, or competencies rather than the internal objects produced by a predictive early-warning pipeline.
	
Provenance modeling provides the conceptual and technical basis for connecting analytic outputs to their sources. The W3C PROV model represents entities, activities, agents, and derivation relations, allowing data products to be linked to the processes and sources from which they were generated~\cite{lebo2013prov,moreau2010}. SHACL complements this perspective by providing a standard mechanism for validating structural constraints in RDF graphs~\cite{knublauch2017shacl}. In early-warning learning analytics, however, provenance is often associated with datasets or processing workflows, while the explanatory structure of individual alerts remains weakly represented. Existing approaches rarely integrate learner states, predictive features, attributions, alerts, evidence, and raw records into a single representation that can be jointly queried and traced.
	
\section{Methodology}
\label{sec:method}

This section presents the incremental Learner Digital Twin for evidence-grounded early warning. We define the temporal prediction problem, introduce the checkpoint-indexed graph, and describe the ontology, synchronization, feature extraction, prediction, attribution mapping, and evidence grounding.

%This section presents the incremental Learner Digital Twin for evidence-grounded early warning. We define the temporal prediction problem, introduce the checkpoint-indexed graph representation, and describe the ontology, synchronization process, feature extraction, prediction, attribution mapping, and evidence-grounding mechanism.

\begingroup
\setlength{\abovedisplayskip}{2pt}
\setlength{\belowdisplayskip}{2pt}
\setlength{\abovedisplayshortskip}{1pt}
\setlength{\belowdisplayshortskip}{1pt}

\subsection{Problem Formulation}
\label{sec:method:problem}

We consider an early-warning setting in which predictions are produced at several moments during a course presentation, before the final learner outcome is known. Let $\mathcal{L}$ denote the set of learners enrolled in a given presentation, and let $D$ denote the set of timestamped raw records produced by the learning platform. These records may include learner--resource interaction records, assessment-result records, registration events, and other platform events relevant to learner activity. Each record $d \in D$ is associated with a timestamp $t(d)$, expressed in the temporal unit used by the platform, such as days or weeks relative to the beginning of the presentation.

Given an ordered sequence of temporal checkpoints $\tau_1 < \tau_2 < \dots < \tau_n$, the data available at checkpoint $\tau$ are defined as
\begin{equation}
	D_{\le \tau} = \{ d \in D \mid t(d) < \tau \}.
	\label{eq:available-data}
\end{equation}
The strict inequality enforces the temporal constraint of the early-warning task: at checkpoint $\tau$, the system may only use records that would have been observed before that time. This prevents information from future interactions, future assessment results, or final outcomes from leaking into the prediction process.

For each learner $\ell \in \mathcal{L}$, the target label is defined from the final course outcome:
\begin{equation}
	y_{\ell} =
	\begin{cases}
		1, & \text{if learner } \ell \text{ has an unsuccessful final outcome},\\
		0, & \text{if learner } \ell \text{ has a successful final outcome}.
	\end{cases}
	\label{eq:label}
\end{equation}
The precise mapping between institutional outcome categories and the binary label is dataset-dependent and is specified in the experimental setup. Although $y_{\ell}$ is used for training and evaluation, it is not available to the model at prediction time. At checkpoint $\tau_i$, the model must therefore estimate the risk of learner $\ell$ using only the temporally bounded data $D_{\le \tau_i}$.

Let $G_{\tau_i}$ denote the \ldt{} graph constructed from the records available before checkpoint $\tau_i$, as defined in Sect.~\ref{sec:method:overview}. For each active learner $\ell$ at $\tau_i$, a feature-extraction function derives a checkpoint-specific feature vector from the graph:
\begin{equation}
	x_{\ell,\tau_i} = \psi_{\mathcal{Q}}(G_{\tau_i}, \ell),
	\label{eq:feature-function}
\end{equation}
where $\psi_{\mathcal{Q}}$ denotes the feature-extraction process implemented by the SPARQL query catalogue $\mathcal{Q}$. The early-warning model is a predictive function
\begin{equation}
	\hat f_{\theta}: \mathbb{R}^{p} \rightarrow [0,1],
	\label{eq:prediction-function}
\end{equation}
which maps $x_{\ell,\tau_i}$ to a risk probability:
\begin{equation}
	\hat p_{\ell,\tau_i} = \hat f_{\theta}(x_{\ell,\tau_i}).
	\label{eq:risk-score}
\end{equation}
A risk alert is emitted when the estimated probability exceeds a decision threshold $\gamma$:
\begin{equation}
	a_{\ell,\tau_i} =
	\mathbb{I}\big[\hat p_{\ell,\tau_i} \geq \gamma\big],
	\label{eq:alert-rule}
\end{equation}
where $a_{\ell,\tau_i}=1$ denotes that an alert is generated for learner $\ell$ at checkpoint $\tau_i$.

The problem addressed in this paper is not only to estimate $\hat p_{\ell,\tau_i}$, but also to preserve the evidential basis of the prediction. For each generated alert, the system should represent the predictive features used by $\hat f_{\theta}$, their feature-level attributions, the evidence supporting these features, and the raw records from which the evidence was derived. Formally, for an attributed feature $f_j$, the explanation is considered evidence-grounded only if there exists at least one evidence item $e$ and one raw record $r$ such that
\begin{equation}
	\textit{supportedBy}(f_j,e)
	\;\wedge\;
	\textit{derivedFrom}(e,r)
	\;\wedge\;
	\textit{RawRecord}(r).
	\label{eq:evidence-grounding}
\end{equation}
Thus, the methodological objective is twofold: first, to construct temporally valid predictive features from the learner state available at each checkpoint; second, to maintain explicit, queryable links from alerts to attributions, from attributions to features, from features to evidence, and from evidence to raw platform records.

Consider, for example, the temporal data structure illustrated in the first stage of Fig.~\ref{fig:method-overview}. At checkpoint $\tau_i$ corresponding to day~25, only records observed before this day are available for analysis. These may include earlier interaction traces, registration information, and assessment results already recorded before the checkpoint. By contrast, records occurring after day~25, future assessment results, and the final course outcome are excluded from prediction and explanation. The learner representation at $\tau_i$ must therefore reflect only what the platform would have known at that moment. This makes the problem both predictive and evidential: the system must estimate learner risk while preserving traceable links to the data available at prediction time.

\begin{figure}[t]
	\centering
	\includegraphics[width=0.98\textwidth]{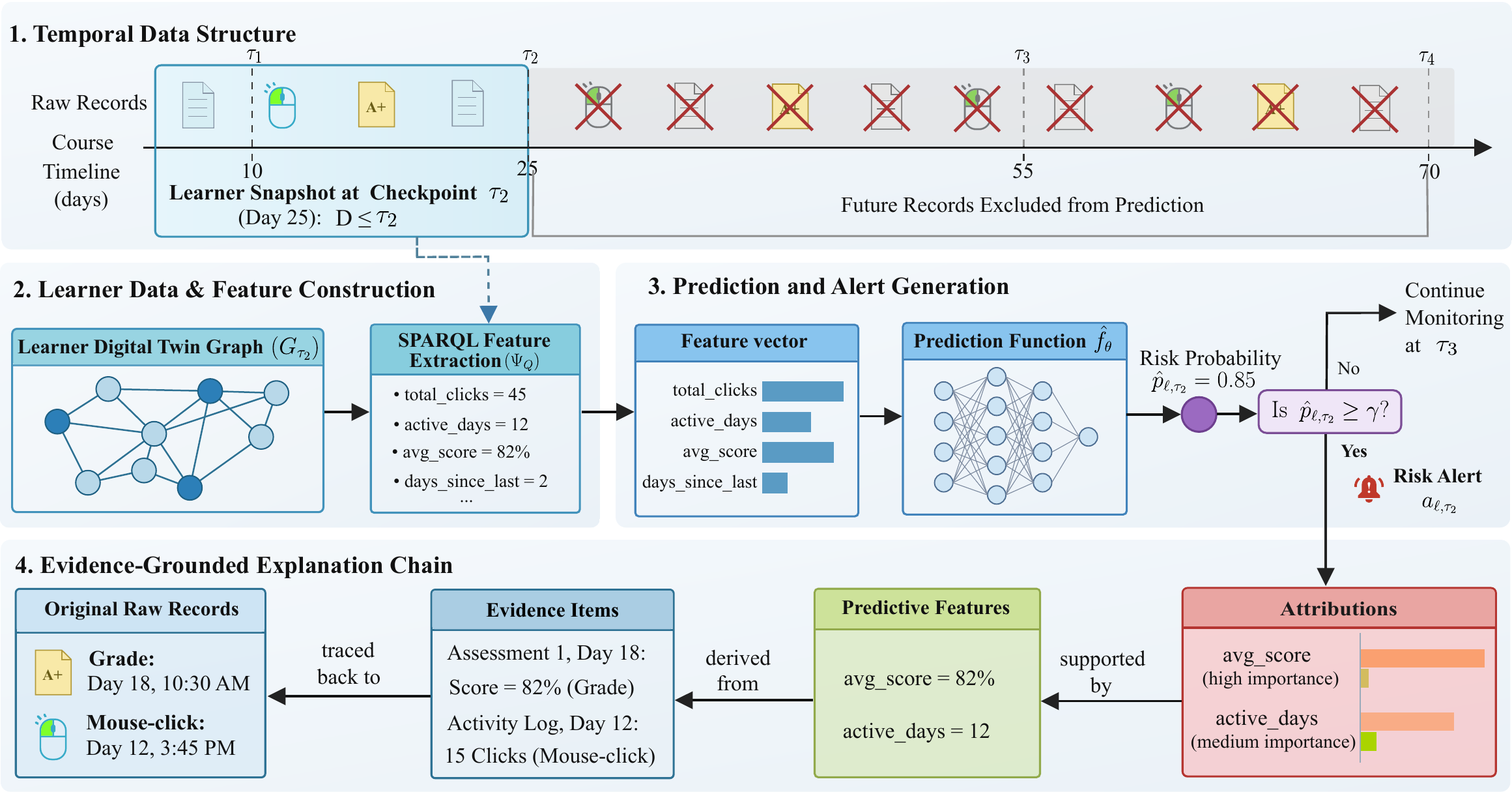}
	\vspace{-0.5cm}
	\caption{Generic workflow of the incremental \ldt{} for evidence-grounded early warning.}
	\label{fig:method-overview}
\end{figure}

\subsection{Overview of the Incremental Learner Digital Twin}
\label{sec:method:overview}

\begin{definition}[Learner Digital Twin]
	A \emph{Learner Digital Twin} is a time-indexed digital representation of a learner, synchronized from observed learning-platform data to support analysis, prediction, explanation, and traceability. Following the general view of digital twins as data-driven representations connected to observed systems~\cite{grieves2017,tao2018digital,tao2019dt}, and their adaptation to educational settings~\cite{habarurema2025enhancing}, this paper operationalizes the learner twin as a checkpoint-indexed knowledge graph $G_{\tau_i}$.
\end{definition}

Building on this definition, the \ldt{} represents, at each temporal checkpoint $\tau_i$, both the learner state derived from temporally available platform records and the analytic artifacts generated by the early-warning pipeline. The Learner Digital Twin at checkpoint $\tau_i$ is defined as
\begin{equation}
	G_{\tau_i} =
	\underbrace{\mu\big(D_{\le \tau_i}\big)}_{\text{observation layer}}
	\;\cup\;
	\underbrace{\Phi_{\tau_1} \cup \dots \cup \Phi_{\tau_i}}_{\text{analytic layer}},
	\label{eq:twin}
\end{equation}
where $\mu$ is a declarative mapping from raw platform records to RDF triples over the \ldt{} ontology, and $\Phi_{\tau_j}$ denotes the set of analytic triples materialized at checkpoint $\tau_j$. The observation layer captures the learner state available before $\tau_i$, while the analytic layer stores the objects produced by the early-warning pipeline, including prediction features, feature attributions, risk alerts, evidence items, and prediction-run metadata.

The observation layer grows monotonically as the course progresses, so that
($\mu(D_{\le \tau_i}) \subseteq \mu(D_{\le \tau_{i+1}})$). The analytic layer is checkpoint-versioned: features, attributions, and alerts generated at an earlier checkpoint are not overwritten by later computations, but remain linked to the corresponding \textit{Checkpoint} entity. This design makes it possible to query the analytic history of the twin, for example to determine which learner state, feature values, and evidence items were available when a specific alert was generated.

Figure~\ref{fig:method-overview} summarizes the workflow implemented at each checkpoint. Temporally available records are first mapped into the \ldt{} graph. The graph is then queried to construct checkpoint-specific feature vectors, which are passed to the predictive function. If the estimated risk exceeds the alert threshold, a risk alert is generated. The alert is then linked to feature attributions, predictive features, evidence items, and the raw records from which the evidence was derived. Thus, the same graph-based representation supports both prediction and traceability.

\subsection{Ontology and Evidence Path}
\label{sec:method:onto}

The \ldt{} ontology is designed as a lightweight and task-oriented model for evidence-grounded early warning. It does not aim to cover all aspects of learner modeling or educational knowledge representation. Instead, it contains the classes and properties required to represent the entities involved in the early-warning workflow: platform observations, predictive features, attributions, alerts, evidence items, checkpoints, and raw records. Consistent with Eq.~\eqref{eq:twin}, the ontology is organized into an observation layer and an analytic layer.

The observation layer represents entities directly derived from platform data, including \textit{Learner}, \textit{Course}, \textit{LearningResource}, \textit{LearningTrace}, \textit{Assessment}, \textit{AssessmentResult}, and \textit{RegistrationEvent}. Each observation-layer entity derived from a platform row is linked to a \textit{RawRecord} individual through \textit{ldto:fromRecord}. The \textit{RawRecord} class stores the source table or stream name and record identifier, thereby anchoring the graph in the original data.

The analytic layer represents the objects produced by the early-warning pipeline. It includes \textit{PredictionFeature}, \textit{FeatureAttribution}, \textit{RiskAlert}, \textit{EvidenceItem}, \textit{PredictionRun}, and \textit{Checkpoint}. A \textit{PredictionFeature} represents one feature value for one learner at one checkpoint and is linked to the evidence items that support its computation. A \textit{FeatureAttribution} stores the contribution of a feature to a specific alert. A \textit{RiskAlert} represents the generated risk decision and is linked to the prediction run, the corresponding checkpoint, the relevant attributions, and the evidence selected for presentation.

\begin{figure}[t]
	\centering
	\includegraphics[width=0.91\textwidth]{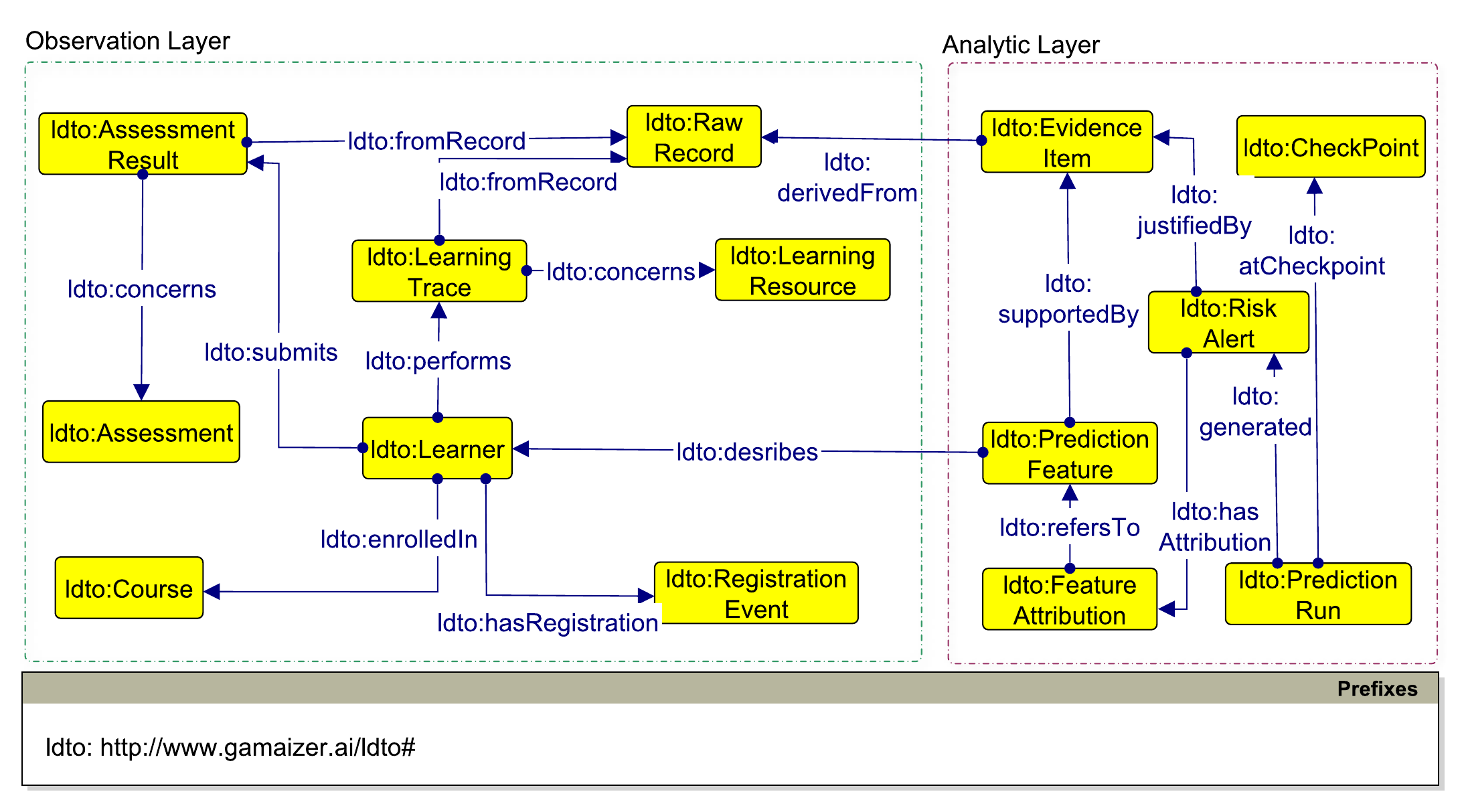}
	\vspace{-0.5cm}
	\caption{Core \ldt{} ontology: observation layer, analytic layer,  evidence links.}
	\label{fig:onto}
\end{figure}

An \textit{EvidenceItem} represents a typed and human-readable statement supporting a predictive feature. It may correspond to activity evidence, assessment evidence, temporal evidence, absence evidence, or another evidence type defined by the application. Each evidence item carries a textual description, a strength value, a temporal scope, and one or more links to \textit{RawRecord} individuals through \textit{ldto:derivedFrom}. The ontology aligns selected classes and properties with PROV-O: evidence items, raw records, prediction features, feature attributions, and alerts are modeled as provenance entities, while prediction runs are modeled as provenance activities.

The explanation of a generated alert is represented as the following graph path:
\begin{equation}
	\begin{aligned}
		\mathit{RiskAlert} 
		&\xrightarrow{\textit{hasAttribution}} \mathit{FeatureAttribution}
		\xrightarrow{\textit{refersTo}} \mathit{PredictionFeature} \\
		&\xrightarrow{\textit{supportedBy}} \mathit{EvidenceItem}
		\xrightarrow{\textit{derivedFrom}} \mathit{RawRecord}.
	\end{aligned}
	\label{eq:chain}
\end{equation}
This path is the central evidence-grounding structure of the KG-based \ldt{}. It makes the explanation chain queryable, verifiable, and traceable through standard graph queries.

\subsection{Synchronization, Feature Extraction, and Alert Grounding}
\label{sec:method:sync}

At each checkpoint $\tau_i$, the replay scheduler exposes only records available before the checkpoint to the RDF mapping process. The set of newly available records between two consecutive checkpoints is denoted by
\begin{equation}
	\Delta_i = D_{\le \tau_i} \setminus D_{\le \tau_{i-1}}.
	\label{eq:delta-records}
\end{equation}
Only records in $\Delta_i$ are added to the observation layer during the update from $G_{\tau_{i-1}}$ to $G_{\tau_i}$. Assessment-related entities are included only when their timestamp or due date satisfies the temporal constraint. This prevents future assessment information from entering the learner state or the predictive features.

The graph is extended with newly available observation-layer triples, then validated against SHACL shapes before analytic write-back. 
The shapes encode the structural conditions required by the explanation mechanism: each \textit{LearningTrace} must be linked to a learner, resource, timestamp, and \textit{RawRecord}; each \textit{PredictionFeature} must have a learner, checkpoint, name, and value; each \textit{EvidenceItem} must be derived from at least one \textit{RawRecord}; and each \textit{RiskAlert} must be generated by one \textit{PredictionRun} and justified by evidence.

%The shapes encode the structural conditions required by the explanation mechanism: each \textit{LearningTrace} must be linked to a learner and a resource; each \textit{EvidenceItem} must be derived from at least one \textit{RawRecord}; and each \textit{RiskAlert} must be generated by one \textit{PredictionRun} and justified by at least one evidence item.

Algorithm~\ref{alg:update} summarizes the checkpoint update process. Feature values are extracted from the graph, materialized as \textit{PredictionFeature} entities, and linked to supporting evidence. The predictive model scores each active learner, and attributions, alerts, and evidence links are written back into the analytic layer. The output of one checkpoint becomes the input graph for the next checkpoint.

\begin{algorithm}[t]
	\setlength{\abovecaptionskip}{2pt}
	\setlength{\belowcaptionskip}{2pt}
	\caption{Checkpoint update of the Learner Digital Twin}
	\label{alg:update}
	\scriptsize
	\begin{algorithmic}[1]
		\setlength{\itemsep}{0pt}
		\setlength{\parskip}{0pt}
		\Require twin $G_{\tau_{i-1}}$, new records $\Delta_i$, shapes $\mathcal{S}$, trained classifier $\hat f_{\theta}$, feature queries $\mathcal{Q}$, threshold $\gamma$
		\State $G \gets G_{\tau_{i-1}} \cup \mu(\Delta_i)$
		\State \textbf{assert} $\mathrm{SHACL}(G,\mathcal{S}) = \emptyset$
		\ForAll{active learner $\ell$ at $\tau_i$}
		\State $x_{\ell,\tau_i} \gets \psi_{\mathcal{Q}}(G,\ell)$
		\State materialize \textit{PredictionFeature} nodes and evidence links
		\State $\hat p_{\ell,\tau_i} \gets \hat f_\theta(x_{\ell,\tau_i})$
		\State $\phi_{\ell,\tau_i} \gets \mathrm{Explain}(\hat f_\theta,x_{\ell,\tau_i})$
		\If{$\hat p_{\ell,\tau_i} \geq \gamma$}
		\State materialize \textit{RiskAlert} and \textit{FeatureAttribution}s
		\State select and link presented evidence through \textit{justifiedBy}
		\EndIf
		\EndFor
		\State \textbf{assert} $\mathrm{SHACL}(G,\mathcal{S}) = \emptyset$
		\State \Return $G_{\tau_i}$
	\end{algorithmic}
\end{algorithm}

Overall, the methodology defines a checkpoint-indexed \ldt{} in which temporally available records are mapped to an observation graph, predictive features are extracted from this graph, and generated alerts are linked back to feature attributions, evidence items, and raw records through the evidence path in Eq.~\eqref{eq:chain}. The next section describes how this framework is instantiated and evaluated on a public learning-analytics benchmark.

\section{Experimentation}
\label{sec:exp}

Our experimental study addresses three research questions: \emph{RQ1} whether the \ldt{} can be incrementally constructed and validated at temporal checkpoints; \emph{RQ2} whether graph-mediated feature extraction preserves the predictive behavior of an equivalent raw-table feature pipeline; and \emph{RQ3} whether generated alerts are grounded in traceable evidence chains.

The experimental instantiation uses a compact feature catalogue covering activity indicators, resource-use counts, assessment progress, temporal behavior, and absence-based signals such as no activity or no submitted assessment before a checkpoint. The same feature definitions are used in both experimental variants.

%The experimental instantiation uses a feature catalogue covering activity, resource use, assessment, temporal behavior, and absence-based indicators. The raw-table and KG-mediated variants compute the same feature definitions; only the extraction substrate and provenance machinery differ. Consequently, the comparison between raw feature engineering and graph-mediated feature extraction evaluates whether the KG substrate preserves the predictive behavior of the equivalent tabular pipeline.

	\subsection{Dataset and Temporal Protocol}
	
The empirical evaluation uses OULAD~\cite{kuzilek2017} as a public benchmark for early-warning learning analytics. The dataset covers 22 presentations of 7 modules, 32{,}593 students, and about 10.6 million VLE interaction records. Its day-based temporal structure and its inclusion of registrations, assessments, and final outcomes make it suitable for early-warning prediction and evidence-grounded alert generation.

Following common practice, learners who withdrew before the presentation start (\textit{date\_unregistration} $\leq 0$) are excluded, since no in-course behavior exists to twin. Learners with no VLE activity and no assessment record before the first checkpoint are retained; their emptiness is itself predictive and generates absence evidence. After preprocessing, six modules have at least one 2013 presentation for training and one 2014 presentation for testing. Module \textit{CCC} is excluded because it has no 2013 presentation in the filtered protocol.
	
We adopt a presentation-based temporal protocol: for each module, the classifier is trained on the 2013 presentations and tested on the 2014 presentations of the same module, simulating deployment on the next cohort. The checkpoints are day 28 (Week~4), day 56 (Week~8), and day 84 (Week~12). Final outcomes are binarized by mapping \textit{Fail} and \textit{Withdrawn} to the at-risk class, and \textit{Pass} and \textit{Distinction} to the non-risk class.
	
	\begin{table}[t]
		\centering
		\caption{OULAD protocol and ontology-mapping summary.}
		\vspace{-0.2cm}
		\label{tab:protocol}
		\scriptsize
		\begin{tabular}{@{}l l@{}}
			\toprule
			Element & Setting \\
			\midrule
			Dataset & OULAD; six eligible modules after preprocessing \\
			Train/test & 2013 presentations for training; 2014 presentations for testing \\
			Checkpoints & Day 28 (Week 4), day 56 (Week 8), day 84 (Week 12) \\
			Target & Fail/Withdrawn = at-risk; Pass/Distinction = non-risk \\
			Observation tables & \textit{studentVle}, \textit{studentAssessment}, \textit{studentRegistration} \\
			Mapped observation entities & \textit{LearningTrace}, \textit{AssessmentResult}, \textit{RegistrationEvent}, \textit{RawRecord} \\
			Analytic entities & \textit{PredictionFeature}, \textit{FeatureAttribution}, \textit{RiskAlert}, \textit{EvidenceItem} \\
			Evaluation & Paired raw-table/KG features; mean over six modules \\
			\bottomrule
		\end{tabular}
	\end{table}
	
\subsection{Baselines and Evaluation Metrics}
\label{sec:setup:baselines-metrics}

The experiment evaluates seven paired model families at each checkpoint: XGBoost (XGB)~\cite{chen2016xgboost}, Random Forest (RF)~\cite{breiman2001random}, Extra Trees (ET)~\cite{geurts2006extremely}, HistGradientBoosting (HGB)~\cite{friedman2001greedy}, LightGBM (LGBM)~\cite{ke2017lightgbm}, CatBoost (CAT)~\cite{prokhorenkova2018catboost}, and a numeric TabTransformer-style neural model (TABTX)~\cite{huang2020tabtransformer}. Each family is run twice: once on conventional raw-table features and once on knowledge-graph-mediated features extracted from the twin. The ontology, SHACL shapes, SPARQL queries, scripts, and per-module results are publicly available\footnote{\url{https://github.com/lengocluyen/ldt-early-warning}}.

The conventional and KG-mediated catalogues compute the same feature definitions; what differs is the substrate and the provenance machinery. XGB-\ldt{} is used as the alert-materialization configuration for the explanation-grounding analysis because TreeSHAP attributions are stable, efficient, and naturally aligned with the feature-attribution graph~\cite{lundberg2018consistent}. %Unless otherwise stated, alerts are generated with threshold $\gamma=0.5$.
Unless otherwise stated, alerts are generated with a fixed threshold  $\gamma=0.5$ to evaluate evidence materialization under a simple reproducible alert policy; deployment-specific threshold calibration is left for future work.

For predictive performance, we compute precision, recall, F1 on the at-risk class, and AUC at each checkpoint. To keep the main benchmark table compact, Table~\ref{tab:predictive} reports F1 and AUC, while precision and recall are retained in the public  artifact. The success criterion is parity: for each paired model family, the KG-mediated variant should match the corresponding raw-table variant when both use the same feature definitions and the same tuning budget.

For incremental construction, we report the number of updated learners, newly integrated traces and assessment results, newly added triples, cumulative graph size, and SHACL violations~\cite{knublauch2017shacl}. For explanation grounding, we define three task-specific metrics over the generated alerts $\mathcal{A}$. Let $T_k(a)$ be the top-(k) attributed features for alert $a$, $E(f)$ the evidence items linked to feature $f$, and $R(e)$ the raw records linked to evidence item $e$ through \textit{derivedFrom}~\cite{lebo2013prov}. Evidence $Coverage@(k)$ measures whether top-(k) features are supported by evidence reaching raw records:
\begin{equation}
	\mathrm{Cov@}k =
	\frac{1}{|\mathcal{A}|}
	\sum_{a \in \mathcal{A}}
	\frac{1}{k}
	\sum_{f \in T_k(a)}
	\mathbb{I}\big[\exists e \in E(f), \exists r \in R(e)\big].
	\label{eq:coverage}
\end{equation}
Presented $Fidelity@(k)$ measures whether presented evidence supports features among the top-(k) model drivers:
\begin{equation}
	\mathrm{Fid@}k =
	\frac{1}{|\mathcal{A}|}
	\sum_{a \in \mathcal{A}}
	\frac{|P(a) \cap T_k(a)|}{|P(a)|},
	\label{eq:fidelity}
\end{equation}
where $P(a)$ is the set of features supported by the evidence presented for alert $a$. Traceability is the fraction of evidence items linked to at least one raw record:
\begin{equation}
	\mathrm{Traceability} =
	\frac{|\{e \in \mathcal{E} \mid \exists r \in R(e)\}|}{|\mathcal{E}|}.
	\label{eq:traceability}
\end{equation}
We report direct coverage for feature-specific activity or assessment records, and contextual coverage for absence or inactivity features grounded in records.

\endgroup
	
	\subsection{Results}
	
	All results reported in this section are obtained from the OULAD instantiation described above.
	
	\vspace{-0.3cm}
	\paragraph{RQ1 - Incremental construction and validation:}
	Table~\ref{tab:construction} reports the incremental construction results, averaged over the six test modules. The test graph contains 2{,}210 learners on average per module. The number of newly integrated traces decreases over time because a large share of the available activity occurs before the first checkpoint: the update adds on average 148{,}560 traces at Week~4, 106{,}204 at Week~8, and 66{,}638 at Week~12. The persistent graph nevertheless grows monotonically, reaching 321{,}401 cumulative learning traces and 3.28M triples on average by Week~12. Both observation graphs and analytic graphs conform to the SHACL shapes at all checkpoints, with zero violations.
	
	\begin{table}[t]
		\centering
		\caption{Incremental twin construction per checkpoint; mean over 6 test modules.}
		\vspace{-0.3cm}
		\label{tab:construction}
		\scriptsize
		\begin{tabular}{@{}lrrrrrrr@{}}
			\toprule
			Checkpoint & Learners & New traces & New results & New triples &
			Cum.\ traces & Cum.\ triples & SHACL viol. \\
			\midrule
			Week 4  & 2{,}210 & 148{,}560 & 1{,}386 & 1{,}349{,}804 & 148{,}560 & 1{,}367{,}486 & 0 \\
			Week 8  & 2{,}210 & 106{,}204 & 1{,}674 &   970{,}948 & 254{,}763 & 2{,}523{,}586 & 0 \\
			Week 12 & 2{,}210 &  66{,}638 &   950 &   608{,}326 & 321{,}401 & 3{,}278{,}846 & 0 \\
			\bottomrule
		\end{tabular}
	\end{table}
	
	These results show that the twin can be maintained incrementally while preserving the structural invariants required for traceable explanations. The graph grows as expected with newly available learning traces and assessment results, while SHACL validation confirms that generated evidence chains remain structurally valid.

\vspace{-0.3cm}
\paragraph{RQ2 - Predictive parity:}
Table~\ref{tab:predictive} reports the KG-mediated early-warning performance on OULAD. The corresponding raw-table results are omitted for compactness because they are identical to the KG-mediated results for every model, checkpoint, and metric. Performance improves from Week~4 to Week~12 as more activity and assessment evidence becomes available. The strongest KG-mediated model is TABTX at Week~4 (AUC 0.735), XGB at Week~8 (AUC 0.809), and ET at Week~12 (AUC 0.847). The central result is therefore feature-substrate parity: graph-mediated feature extraction preserves the predictive behavior of the equivalent raw-table pipeline while adding provenance, validation, and evidence materialization.

\vspace{-0.5cm}
\begin{table}[h]
	\centering
	\caption{KG-mediated early-warning performance over six modules; raw-table results are identical and omitted for compactness.}
	\vspace{-0.3cm}
	\label{tab:predictive}
	\scriptsize
	\begin{tabular}{lrrrrrr}
		\toprule
		& \multicolumn{2}{c}{Week 4} & \multicolumn{2}{c}{Week 8} & \multicolumn{2}{c}{Week 12} \\
		\cmidrule(lr){2-3}\cmidrule(lr){4-5}\cmidrule(lr){6-7}
		Method & KG F1 & KG AUC & KG F1 & KG AUC & KG F1 & KG AUC \\
		\midrule
		XGB   & 0.610 & 0.735 & 0.680 & 0.809 & 0.718 & 0.842 \\
		RF    & 0.597 & 0.725 & 0.653 & 0.795 & 0.719 & 0.840 \\
		ET    & 0.605 & 0.733 & 0.674 & 0.805 & 0.723 & 0.847 \\
		HGB   & 0.581 & 0.703 & 0.644 & 0.778 & 0.710 & 0.822 \\
		LGBM  & 0.588 & 0.700 & 0.645 & 0.774 & 0.702 & 0.815 \\
		CAT   & 0.600 & 0.716 & 0.661 & 0.787 & 0.715 & 0.834 \\
		TABTX & 0.619 & 0.735 & 0.658 & 0.785 & 0.701 & 0.821 \\
		\bottomrule
	\end{tabular}
	\vspace{-0.5cm}
\end{table}

\vspace{-0.6cm}
\paragraph{RQ3 - Evidence grounding:}
Table~\ref{tab:grounding} reports the explanation-grounding metrics for the fixed XGB-\ldt{} alert generator on OULAD. $Direct@3$ and $Direct@5$ measure whether the top attributed features are supported by evidence chains reaching feature-specific activity or assessment records. Contextual coverage and traceability are equal to 1.0 at all checkpoints, showing that every generated evidence item is linked to a traceable graph basis. Direct coverage is lower because absence and inactivity features are grounded through contextual records, such as registrations or due-assessment metadata, rather than feature-specific activity rows. $Fid@3$ decreases at later checkpoints as the diversity-aware evidence policy selects a broader set of assessment and temporal evidence.

\vspace{-0.5cm}
\begin{table}[h]
	\centering
	\caption{Explanation-grounding metrics for the fixed XGB-LDT alert generator; alerts are aggregated over the six test modules.}
\vspace{-0.3cm}
	\label{tab:grounding}
	\scriptsize
	\begin{tabular}{@{}lrrrr@{}}
		\toprule
		Checkpoint & Alerts & Direct@3 & Direct@5 & Fid@3 \\
		\midrule
		Week 4  & 5{,}914 & $0.764\pm0.119$ & $0.760\pm0.120$ & $0.898\pm0.069$ \\
		Week 8  & 5{,}609 & $0.706\pm0.126$ & $0.695\pm0.088$ & $0.828\pm0.131$ \\
		Week 12 & 6{,}385 & $0.648\pm0.150$ & $0.665\pm0.079$ & $0.744\pm0.063$ \\
		\bottomrule
		\vspace{-0.2cm}
	\end{tabular}
	\vspace{2pt}
	\parbox{0.92\linewidth}{\footnotesize \textit{Note.} $Direct@k$ is direct evidence coverage for the top-$k$ attributed features; $Fid@k$ denotes presented fidelity. Contextual $Cov@3$=Contextual $Cov@5$=1.000, traceability=1.000, and evidence items per alert=5.0 at every checkpoint.}
\vspace{-1.0cm}
\end{table}

\paragraph{Case-level evidence:}
We close with two Week~8 alerts generated by the XGB-\ldt{} run for the AAA/2014J test presentation. Learner identifiers are the anonymized OULAD identifiers. The first case is a true positive. The model assigns learner \#111512 a risk probability of 0.928, and the learner's final outcome in OULAD is \emph{Fail}. The instructor-facing explanation reads:

\vspace{-0.2cm}
	\begin{quote}\footnotesize\itshape
		Learner \#111512 is predicted to be at risk at week 8 because: (1) no assessment due before day 56 had been submitted, so both \texttt{n\_submitted} and \texttt{avg\_score} are 0; (2) two TMAs were already due, on days 19 and 54; and (3) the learner's last VLE activity before the checkpoint was on day 34, giving \texttt{days\_since\_last}=22 and an activity change from 25 clicks in days 28--41 to 0 clicks in days 42--55.
	\end{quote}\vspace{-0.2cm}

The second case is a borderline false positive. The model assigns learner \#1112555 a risk probability of 0.517 at the same checkpoint, while the final outcome is \emph{Pass}. The explanation reads:

\vspace{-0.2cm}
	\begin{quote}\footnotesize\itshape
		Learner \#1112555 is predicted to be at risk at week 8 because: (1) one assessment due before day 56 had not yet been submitted; (2) \texttt{missing\_assessment}=1, although one earlier TMA was submitted on day 19 with a score of 64; and (3) the learner remained recently active, with a VLE interaction on day 55 and \texttt{days\_since\_last}=1.
	\end{quote}
	\vspace{-0.2cm}

This second case is not presented as a predictive success. The missing assessment was submitted on day 57 with a score of 72, but this information is correctly excluded from the day-56 graph. Together, the two cases illustrate how the \ldt{} distinguishes direct activity and submission records from contextual absence evidence, and how each alert can be traced from attributed features to checkpoint-bounded evidence items and their originating records.

\section{Discussion}
\label{sec:discussion}

The results lead to three findings. First, the \ldt{} can be populated incrementally across checkpoints while preserving the structural constraints required for evidence-grounded alerts. The graph grows from Week~4 to Week~12, and SHACL validation reports no violations, indicating that the observation and analytic layers remain coherent over time.
Second, the KG-mediated pipeline preserves the predictive behavior of the raw-table pipeline. The identical F1 and AUC values are expected because both variants use the same feature definitions; the value of the KG is not predictive improvement, but the explicit connection between features, evidence items, and raw records. Third, alerts are traceable through evidence chains. Direct coverage is lower for absence and inactivity features because these signals are grounded through contextual records, such as registrations or due-assessment metadata, rather than direct activity rows.

A key methodological choice is to make the graph load-bearing rather than merely explanatory. If the graph were populated only after prediction, it would serve as an external explanation store and could diverge from the actual preprocessing pipeline. In contrast, the same representation is used here to derive predictive features and to preserve their evidential basis. This design reduces the risk of inconsistency between raw data processing, model input, and presented explanations. It also makes explanation completeness verifiable through SHACL constraints and graph queries.

The use of the term \emph{Learner Digital Twin} should be understood in a restricted operational sense. The study simulates synchronization by chronologically replaying a historical dataset at predefined checkpoints. It does not evaluate live ingestion, out-of-order events, data correction, real-time system integration, or instructor feedback loops. The implementation therefore demonstrates a checkpoint-indexed \ldt{} with recurrent synchronization and analytic history, rather than a fully deployed live educational digital twin.

%The empirical evaluation is limited to OULAD and to a compact feature catalogue. The twin captures behavioral and assessment evidence, but it does not represent competency states, learning objectives, or pedagogical interventions, since such information is not available in OULAD. Future work should evaluate the framework in live learning platforms, extend the ontology toward competency-aware learner states, and study whether evidence-grounded alerts improve instructor understanding, trust, and intervention decisions.

\section{Conclusion}
\label{sec:conclusion}

This paper introduced an incremental knowledge-graph Learner Digital Twin for evidence-grounded early warning in learning analytics. The approach routes prediction through a semantic representation in which learner observations, predictive features, feature attributions, risk alerts, evidence items, and raw records are explicitly represented as graph entities. The \ldt{} therefore serves both as a feature-extraction substrate and as an evidence-grounding substrate.
The empirical instantiation on OULAD shows that graph-mediated feature extraction preserves the predictive behavior of equivalent raw-table features across model families and checkpoints. At the same time, generated alerts can be linked back to feature attributions, evidence items, and raw platform records through queryable evidence chains. These results suggest that semantic provenance can make early-warning pipelines more transparent, traceable, and verifiable without reducing predictive performance.
Future work will extend the learner twin toward competency-aware learner states, evaluate the framework in live learning platforms, and study whether evidence-grounded alerts improve instructors' understanding, trust, and intervention decisions.

\section*{Acknowledgments}
We thank the Ikigai consortium led by the association Games for Citizens, the company Gamaizer, and the FORTEIM project (winner of the AMI CMA France 2030 call for projects), for their support and collaboration. Their contributions have provided significant added value to the completion of this research.

	\bibliographystyle{splncs04}
	\bibliography{references}
	
\end{document}